\documentclass[10pt,twocolumn,letterpaper]{article}

\usepackage{cvpr}      %

\usepackage{soul}
\usepackage{bm}
\usepackage{bbm}
\usepackage{multicol,multirow}
\usepackage{colortbl}
\usepackage{pifont}
\usepackage{overpic}
\definecolor{cvprblue}{rgb}{0.21,0.49,0.74}
\usepackage[pagebackref,breaklinks,colorlinks,allcolors=cvprblue]{hyperref}
\usepackage{textcomp}
\title{Gate-and-Merge: Zero-shot Compositional Personalization\\
of Vision Language Models}

\author{Guodong Ding, Angela Yao\\
National University of Singapore\\
{\tt\small \{dinggd,ayao\}@comp.nus.edu.sg}
}

\begin{document}
\maketitle

\begin{abstract}
This paper tackles compositional personalization of vision–language models (VLMs). In this problem, multiple user-defined concepts must be recognized or described jointly at test time. 
We introduce Gate-and-Merge, a zero-shot framework that enables compositional personalization without the need for co-occurrence training. During personalization, each concept is learned independently as a lightweight LoRA adapter, paired with a concept token. The base model remains unchanged and concepts are kept disentangled. At inference, we enable composition by merging concept-specific LoRA updates directly in weight space. To suppress irrelevant activations and prevent interference, a gating mechanism is employed to estimate textual and visual cues and only select modules contributes to the prediction. We further stabilize composition by combining only the most meaningful and mutually consistent updates, helping preserve each concept’s identity. 
Our quantitative and qualitative analyses show consistent gains in boosting performances on multiple personalization tasks on both the single and the compositional setting.  
\end{abstract}

\section{Introduction}
\label{sec:intro}
Vision Language Models (VLMs)~\cite{li2023blip,liu2023visual,wang2023visionllm} have demonstrated remarkable generalization capabilities across open-world tasks such as recognition, reasoning, and image–text generation. However, they lack the ability to understand personalized concepts, \eg, user-specific faces, pets, artwork styles, or private objects that are absent from large-scale pre-training data. Enabling VLM personalization, \ie, teaching a model to recognize or describe novel, user-defined concepts, is essential for advanced applications in assistive AI, creative content generation, human–robot interaction, and personalized digital assistants.

\begin{figure}[!t]
    \centering
    \begin{overpic}[width=\linewidth]{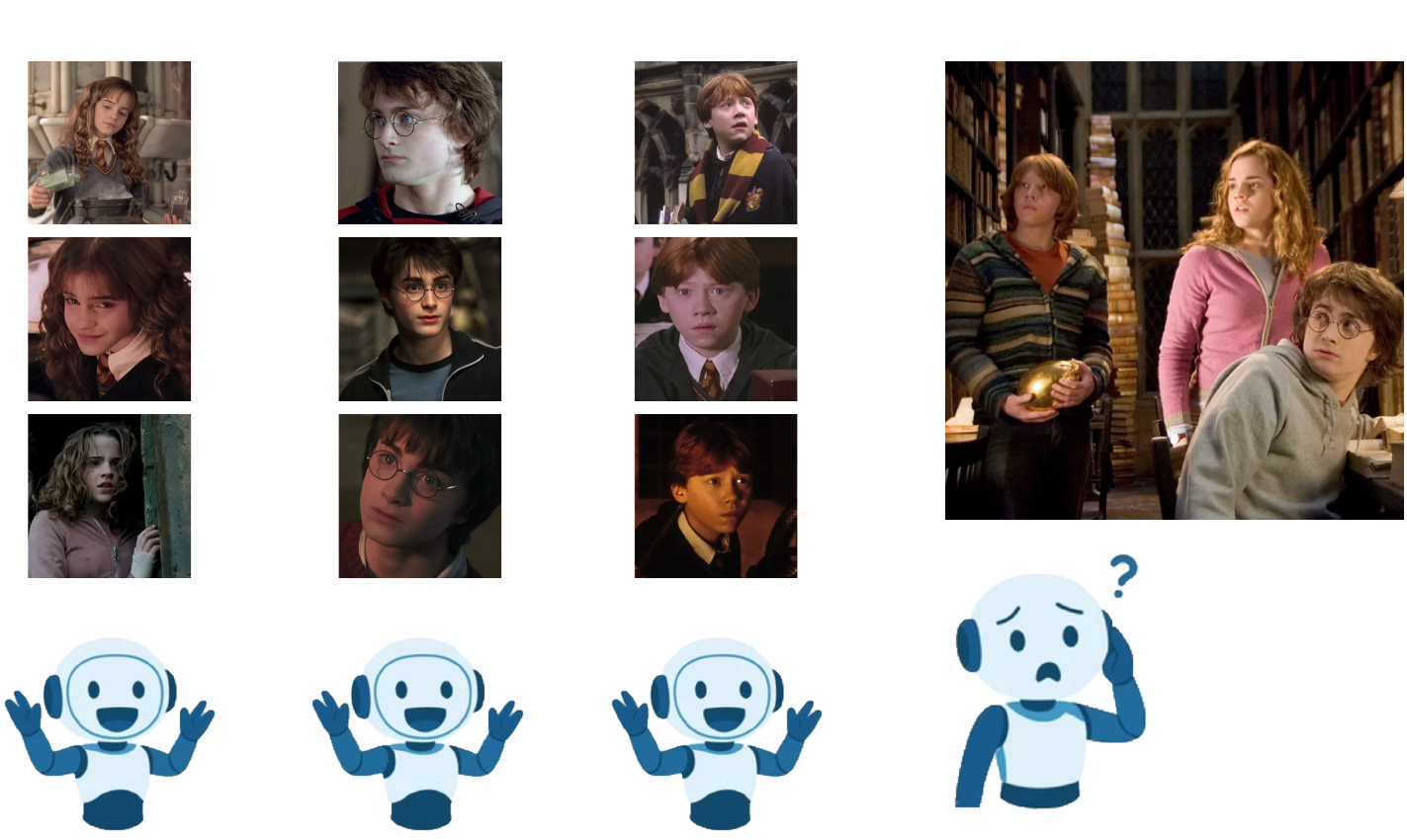}
     \put(5,58){\small \textlangle H\textrangle}
    \put(27,58){\small \textlangle P\textrangle}
    \put(48,58){\small \textlangle R\textrangle}
    \put(0,-4){\small ``She is \textlangle H\textrangle!''}
    \put(23,-4){\small ``He is \textlangle P\textrangle!''}
    \put(45,-4){\small ``He is \textlangle R\textrangle!''}
    
    \put(79,2){\small ``Who are they?''}
    \end{overpic}
    \vspace{0.5em}
    \caption{Personalized VLMs can learn each concept in isolation, but without a mechanism for compositionality, they may struggle to interpret scenes containing multiple user-defined concepts. }
    \label{fig:teaser}
\end{figure}
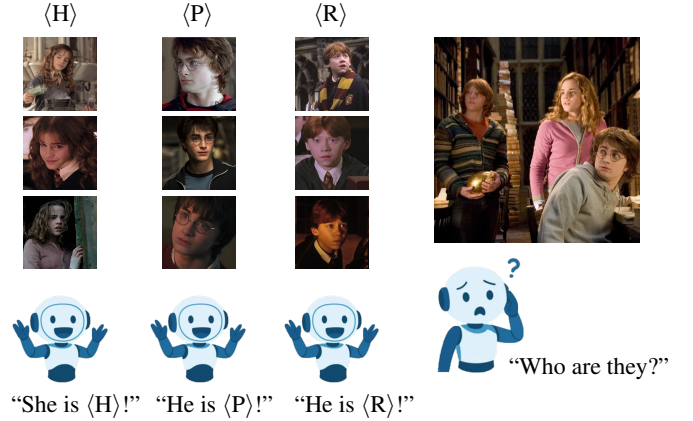

Existing efforts toward VLM personalization can be broadly categorized into parameter-based and retrieval-based approaches. 
Parameter-based methods such as~\cite{alaluf2024myvlm,nguyen2024yo} embed user-specific concepts directly into the model through adaptation techniques such as concept embeddings or learnable prompts. Specifically,~\cite{alaluf2024myvlm} appends detected concept embedding into the vision feature, while~\cite{nguyen2024yo} adopts prompt learning to obtain meaningful concept prompts. These methods typically explored under a \textit{single-concept} setting, where each concept is trained independently. In this regime, the model internalizes each concept in isolation, which limits its ability to recognize or generate compositional instances involving multiple unseen concept combinations, as illsutrated in~\cref{fig:teaser}. Recent extensions have attempted \textit{multi-concept} personalization by joint training of learnable prompt of multiple concepts~\cite{an2024mc}.  Though such training may not require co-occurrence of concepts, it still  entangles concept prompts during learning, making it difficult to isolate and recombine individual concepts in novel configurations.

The retrieval-based approach~\cite{hao2025rap} is an alternative that avoids parameter updates for novel concepts. Instead, it performs instruction tuning on a large corpus containing pseudo-concepts, enabling the model to handle concept-conditioned queries\footnote{We refer to queries containing user-defined concept tokens (\eg, $\langle c\rangle$) as concept-conditioned queries, with concept name $c$ enclosed by $\langle\cdot  \rangle$. One exemplar token can be $\langle my\, cat\rangle$.}. During inference, visual exemplars and relevant information are retrieved from an external database to ground the prediction. While effective, this strategy does not internalize concepts within the model; instead, the knowledge remains external and often depends on storing raw personal data, which raises privacy concerns. Additionally, the fine-tuning strategy in such approach does not explicitly address the composition problem, so it would still face limitations in concept composition. 

Parameter-based concept embedding~\cite{alaluf2024myvlm} and learnable prompts~\cite{nguyen2024yo} have been effective for internalizing user-specific concepts, they face practical limitations in expressiveness, composability, and scalability when handling multiple concepts or requiring modular reuse, as noted by~\cite{wang2023universality}. To overcome these challenges, we propose a module-based approach explicitly designed for scalable and compositional personalization.

In this paper, we propose Gate-and-Merge, a framework for zero-shot compositional personalization of VLMs. Unlike existing methods that rely on concept co-training or external retrieval databases, our approach learns each concept independently using a lightweight LoRA adapter paired with a unique concept token. This modular design allows the model to internalize distinct concept representations that remain disentangled across training instances.  It forms the foundation for flexible composition of multiple concepts at inference time without requiring joint optimization. 

At inference time, we introduce a LoRA merging mechanism that combines the parameter updates of relevant personalized concepts.  The merging, done directly in the weight space, allows the model to compose unseen concept combinations and reason about them zero-shot.  To support the interaction, the gating module identifies which concepts are contextually relevant and selectively activates only the corresponding LoRA adapters.  
Naively summing all LoRA updates introduces interference and has the risk of degrading performance. To mitigate this, we sparsify each LoRA update to keep only the dominant signals, and then merge them in a sign-aware manner so that only aligned updates are combined, avoiding destructive interference.

The contributions of this paper is summarized as follows: 
\textbf{1)} We introduce Gate-and-Merge, a modular framework for zero-shot compositional personalization of VLMs without requiring co-occurrence training or external databases. 
\textbf{2)} We propose to represent each user-defined concept by a lightweight LoRA module and a corresponding concept token, supporting scalable and privacy-preserving personalization. 
\textbf{3)} We introduce a training-free mechanism that selectively activates and merges relevant LoRA modules based on input-dependent textual and visual cues. This enables the model to compose multiple independently learned concepts at inference time.
\textbf{4)} We demonstrate through extensive experiments that our method achieves superior performance compared to parameter-based personalization methods. 

\section{Related work}
\label{sec:related}
\textbf{Personalization of MLLMs.} Personalization aims to adapt a model's knowledge or behavior to user-specific concepts, objects, identities, or preferences. While early personalization efforts~\cite{wei2025personalized,xu2025personalized,zeng2024jedi,ruiz2023dreambooth}, were primarily explored in image generation systems, such as diffusion models. These methods mostly focused on teaching a generative model to render a user’s face, pet, or artwork style. Although these approaches successfully capture visual appearance, they are inherently limited: they personalize only the visual output space, without integrating the concept into reasoning, language, or multimodal understanding.  
The rise of multimodal large language models (MLLMs)~\cite{li2023blip,liu2023visual,wang2023visionllm} has reshaped personalization. Beyond generating customized images, personalization now aims to internalize user-defined concepts so that a model can remember, recognize, and reference them across tasks such as dialogue, VQA, captioning and reasoning.
To pursue this goal, recent methods extend personalization into VLMs. MyVLM~\cite{alaluf2024myvlm} attaches external classification heads for concept recognition and injects learned concept embeddings into the language model to influence responses. Yo'LLaVA~\cite{nguyen2024yo} modifies LLAVA more directly by expanding the tokenizer and learning soft prompts per concept, enabling the model to reference these concepts in text. While effective, these approaches still struggle with scalable, compositional, and multi-concept personalization.

\textbf{Parameter-efficient adaptation.}
Large pretrained models can be efficiently fine-tuned by introducing lightweight, low-rank adapters rather than updating all parameters. Low-Rank Adaptation (LoRA)~\cite{hu2022lora} became one of the most popular methods, inserting trainable rank-decomposition matrices into frozen transformer weights to enable compact, reusable fine-tuning. Subsequent studies~\cite{zhou2022conditional,zhang2022tip,ruiz2023dreambooth} extend this idea to adapters, prompt-tuning, and other lightweight mechanisms across vision and language models. 
While LoRA was initially designed for single-task adaptation, recent work has explored composing multiple LoRA modules, each encoding a distinct skill or concept. 
\cite{tian2024hydralora} proposes an asymmetric PEFT design that shares a common A matrix while learning multiple expert B matrices, using an MoE router to reduce task interference and automatically discover intrinsic data components, outperforming standard and multi-LoRA methods with fewer parameters. 
~\cite{po2024orthogonal} learns LORAs through orthogonal adaptation, enables instant multi-concept diffusion model merging by training each LoRA in orthogonal subspaces, preventing crosswalk and better preserving identities.  
Recently, \cite{prabhakar2025lora} studied skill composition for language models, showing that merging multiple LoRAs trained on disjoint tasks can outperform joint training when data for combined tasks is scarce. Similarly, \cite{roy2025multlfg} proposed Multi-LoRA composition for image generation using frequency-domain guidance to blend multiple concept-specific LoRAs without retraining. \cite{zhou2025compositional} introduced CS-ReFT, which learns orthogonal subspaces for distinct skills and composes them through subspace routing—conceptually parallel to compositional LoRA at the representation level.
Our work extends this line of research by examining LoRA composition in visual concept recognition and multi-concept reasoning. Each LoRA represents a personalized concept or specialization, and the model must dynamically activate and fuse the relevant modules to handle complex, multi-concept queries.

\section{Method}
\label{sec:method}

\subsection{VLM Personalization}
Let $\mathcal{M}$ be a pretrained vision-language model. A user-specific concept $c$ is represented by a textual identifier $t_c$, a small support set of image set $\mathcal{X}_c$, and a text set $\mathcal{L}_c$. The goal of personalization is to obtain a compact concept presentation $\Pi_c$ such that $\mathcal{M}\oplus \Pi_c$ can correctly recognize, describe, or answer questions about $c$ in future image-text inputs.  In existing VLM personalization tasks, $\oplus$ may denote either prompt injection, that prepends learnable prompts to the text encoder input without modifying $\Theta$, or performing parameter update with instruction tuning to teach the model about these concepts.

\begin{figure*}[thb]
\centering
\begin{overpic}[width=1\textwidth]{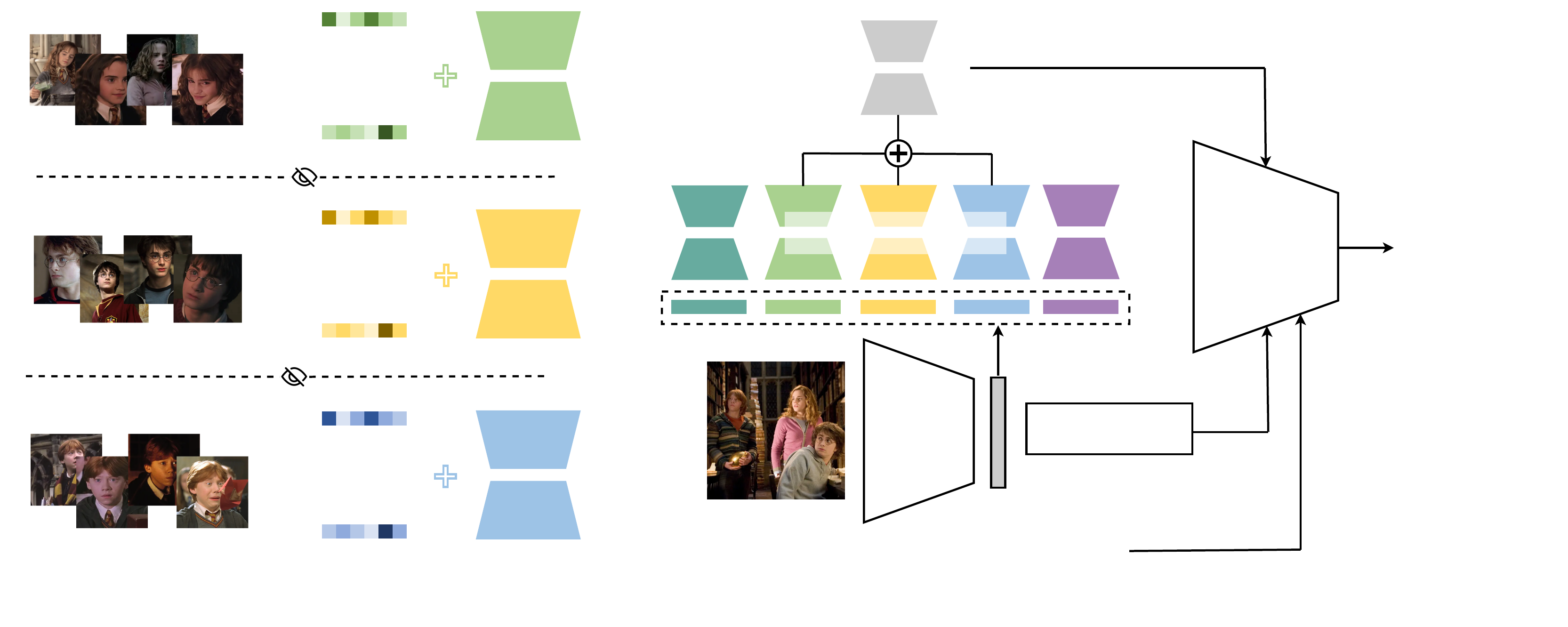}
\put(7,4){\small \textlangle R\textrangle }
\put(7,17){\small \textlangle P\textrangle }
\put(7,30){\small \textlangle H\textrangle }
\put(12,0.5){\small (a) Single concept learning}
\put(62,0.5){\small (b) Multi-concept composition}
\put(23,7){\small $\mathbf{e}$}    
\put(23,19.5){\small $\mathbf{e}$} 
\put(23,32.5){\small $\mathbf{e}$} 
\put(22.5,11.5){\small $\mathbf{w}$} 
\put(22.5,24.5){\small $\mathbf{w}$} 
\put(22.5,36.8){\small $\mathbf{w}$} 

\put(33,37){\small $A$} 
\put(33,32.4){\small $B$} 
\put(33,24){\small $A$} 
\put(33,19.5){\small $B$} 
\put(33,11.5){\small $A$} 
\put(33,7){\small $B$}

\put(37,9.5){\small $\bm{\delta}$} 
\put(37,21.5){\small $\bm{\delta}$} 
\put(37,35){\small $\bm{\delta}$} 

\put(31,3){\small (LoRA)} 
\put(21,3){\small (token)}

\put(56,13){\small Vision }
\put(55.5,10.5){\small Encoder }
\put(66,12){\small Project layer}
\put(77,25){\small Language}
\put(78,22.5){\small Model}
\put(44,4.3){\small ``Please give a caption of the image.'' }

\put(51,24.6){\small (gate and merge)}
\put(55, 39.5){\small $\bm{\delta}_\text{merge}$}
\put(89,24){
            \parbox{0.1\linewidth}{ \small ``From left to right are \textlangle R\textrangle, \textlangle H\textrangle, and \textlangle P\textrangle.'' 
            }}
\end{overpic}
    \caption{The proposed Gate-and-Merge framework. (a) During personalization, each concept is independently learned as a LoRA module associated with a unique concept token, enabling scalable and disentangled representation learning. (b) At inference, the gating function identifies concept relevance from text and image cues, while the merging mechanism fuses the aligned LoRA updates to achieve zero-shot compositional reasoning.}
    \label{fig:method}
\end{figure*}
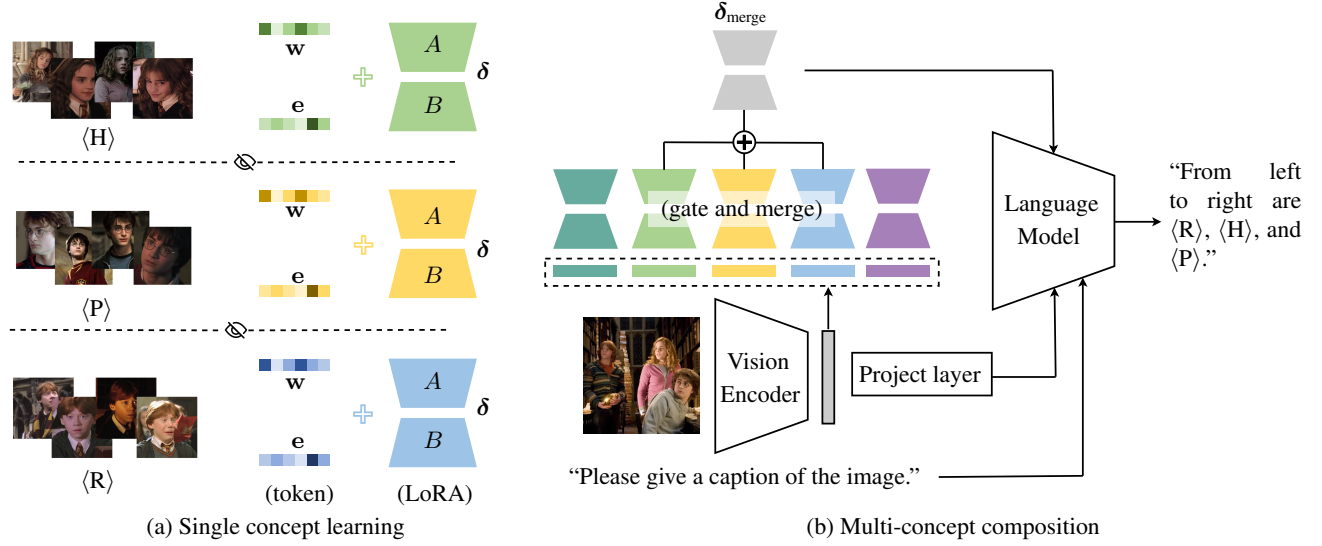

We consider the task of zero-shot compositional personalization of VLMs. Given the pretrained $\mathcal{M}$ and a set of user-defined concepts $\mathcal{C} = \{c_1, c_2, ...,c_K\}$, the goal is to learn compact representation for each concept independently, without observing any co-occurrence of concepts during learning. At inference, the model must correctly recognize, describe or answer questions about novel compositions of these concepts,  \eg, inputs containing multiple concepts from a subset $\mathcal{S}\subseteq \mathcal{C}$, even though no sample in training depicted them together. 
To this end, we propose a Gate-and-Merge framework that can achieve zero-shot compositional personalization. 
Our method consists of two main parts. The first part focuses on single concept learning (\cref{subsec:single}), where user-defined concepts are individually internalized as modular representations. The second part addresses multi-concept composition (\cref{subsec:compo}), enabling the model to selectively activate and fuse the relevant concept modules during inference. The full pipeline is depicted in~\cref{fig:method}.

\subsection{Single Concept Learning}
\label{subsec:single}
Unlike previous work that represents concept information using learnable prompts~\cite{nguyen2024yo}, we propose a more explicitly modular approach. Specifically, our approach models each concept $c$ as an independent module comprising two components: a learnable {concept token} ($\langle c\rangle$), and a set of {low-rank adaptation parameters}. 

\noindent\textbf{Concept Token.} We begin by introducing a new token $\langle c\rangle$ into the tokenizer vocabulary to serve as the textual identifier for concept $c$. This token is associated with a trainable embedding vector, 
\begin{equation}
\mathbf{e}_c = \text{Embedding}(\langle c \rangle) \in \mathbb{R}^d,
\end{equation}
where $d$ is the hidden dimensionality of language model. 
 To enable concept-level decoding, we also extend the model's output head $W_\text{out}\in \mathbb{R}^{V\times d}$ by appending a new row $\mathbf{w}_c$ corresponding to $\langle c\rangle$, yielding:
\begin{equation}
W_{\text{out}} \leftarrow \begin{bmatrix}
W_{\text{out}} \
\mathbf{w}_c^\top
\end{bmatrix}\in \mathbb{R}^{(V+1)\times d}, \quad \mathbf{w}_c \in \mathbb{R}^d.
\end{equation}
Both $\mathbf{e}_c,\mathbf{w}_c$ are randomly initialized and optimized end-to-end during training.

\noindent\textbf{LoRA Parameters.} Beyond token embeddings, we employ concept-specific parameter adaptation $\bm{\delta}_c$ by LoRA. 
Let $W \in\mathbb{R}^{d_\text{out}\times d_\text{in}}$ denote a weight matrix in the language model, we define the concept adaptation as a low-rank residual: 
\begin{equation}
W' = W + \bm{\delta}_c,\,
\bm{\delta}_c = A_c B_c, \,
A_c \in \mathbb{R}^{d_{\text{out}} \times r}, B_c \in \mathbb{R}^{r \times d_\text{in}},
\end{equation}
where $r\ll \min(d_\text{in}, d_\text{out})$ is the rank of the LoRA update. 

In summary, the personalized module for concept $c$ is fully specified by:
\begin{equation}
\Pi_c = \{ \mathbf{e}_c, \mathbf{w}_c, \bm{\delta}_c \},
\end{equation}

Each concept module is trained independently using a small support set $\mathcal{D}$ described above. The goal is to teach the model to correctly associate $\langle c\rangle$  with the visual appearance of the concept. To achieve this, we optimize a standard masked language modeling objective over text, image inputs:
\begin{equation}
\mathcal{L} = - \sum_{t=1}^{T} \log P(\ell_{i,t} \mid x_i,  \ell_{i,<t}; \mathbf{e}_c, \mathbf{w}_c, \bm{\delta}_c),
\end{equation}
where $\ell$ is the target output of length $T$.

During training, adaptation is exclusively to the language branch via concept tokens and LoRA parameters and the vision branch is fixed. Keeping the visual encoder frozen can help maintain a stable and shared feature space across concepts, ensuring that visual features remain mutually compatible and comparable, which is essential for reliably composing multiple independently learned concept modules at inference time. %

\subsection{Multi-concept Composition}
\label{subsec:compo}
 After learning individual concepts as modular components, the model is required to generalize compositionally at inference, \ie, handling inputs that involve multiple concepts at once for zero-shot personalization, despite having no exposure to such combinations during training.
 However, naively summing the LoRA updates for all concepts often causes semantic interference, where different adaptations may conflict and degrade the performance. To overcome this issue, we introduce a selective Gate-and-Merge mechanism that identifies the relevant concepts for a given input and composes only their corresponding LoRA weights, while explicitly mitigate potential interference during the fusion process. %

 \noindent \textbf{Gate.} Formally, given an input image $x$ and textual query $q$, the gate mechanism considers two complementary cues: textual and visual relevance. 
Textual relevance is triggered by the explicit mention of the concept token $\langle c_i\rangle$ in the query:
\begin{equation}
    t_c(q) = \mathbbm{1}[\langle  c_i \rangle \in q],
\end{equation}
while visual relevance is measured via cosine similarity in the feature space. Let $E_v(x) = [\mathbf{v}_1, ..., \mathbf{v}_P]$ be $\ell_2$ normalized patch features from the frozen vision encoder, and let $\phi_c$ be $\ell_2$ normalized prototype for concept $c$. Per-patch cosine scores can be defined as:
\begin{equation}
    s_{j,c} = \langle \mathbf{v}_j, \phi_c\rangle, \quad j=1,...,P.
\end{equation}
and aggregated via top-K pooling:
\begin{equation}
    s_c(x) = s_v^\text{topk}(x,c) = \frac{1}{K}\sum_{j\in \text{topk}\{s_{j,c}(x) \}}s_{j,c}(x)
\end{equation}

A concept is selected if it is either text-mentioned or visually supported:
\begin{equation}\label{eq:gating}
    g_c(x,q) = \mathbbm{1}[t_c(q) =1 \vee s_c(x) \geq \tau],
\end{equation}
where $\tau$ is the pre-defined similarity threshold. The above then yields the set of active concepts:
\begin{equation}
    \mathcal{S} = \{ c\in \mathcal{C}: g_c(x, q)=1\}.
\end{equation}

\noindent\textbf{Merge.} Given the gated set $\mathcal{S}$, we compose only the LoRA updates of the selected concepts. A na\"{i}ve solution is to sum all updates, \ie, $\bm{\hat{\delta}} = \sum_{c\in \mathcal{S}}\bm{\delta}_c$.  However, direct addition may cause interference, because independently trained adapters may update overlapping weight coordinates in different or even opposing directions, creating %
conflicts that distort the intended signals. Instead, we perform a structured merging strategy that combines DARE~\cite{yu2024language} and ties~\cite{yadav2023ties}.  
For each selected LoRA $\bm{\delta}_c$, we first reduce its redundancy in the LoRA parameters by stochastic dropping with unbiased rescaling:
\begin{equation}
\begin{aligned}
\bm{m}_c &\sim \mathrm{Bernoulli}(p),\\
\widetilde{\bm{\delta}}_c &= (\bm{1}-\bm{m}_c)\odot \bm{\delta}_c,\\
\widehat{\bm{\delta}}_c &= \widetilde{\bm{\delta}}_c / (1-p),
\end{aligned}
\label{eq:dare}
\end{equation}
so that $\mathbb{E}[\bm{\hat{\delta}}_c] = \bm{\delta}_c$\footnote{A rough proof can be found in~\cite{yu2024language}}.  In~\cref{eq:dare}, $\bm{m}_c \in \{0,1\}^{\text{shape}(\bm{\delta}_c)}$ is an element-wise Bernoulli mask sampled with the drop rate $p \in[0,1)$. This yields sparse adapters that retain the expected update while suppressing low-contribution parameters, reducing potential cross-concept conflicts. 

Following~\cite{yadav2023ties}, we fuse the sparsified LoRA adapters into a single adapter by enforcing sign consistency at each parameter coordinate and averaging only non-conflicting updates. We index an entry in the weight matrix by \((u,v)\), where \(u\in\{1,\dots,d_{\text{out}}\}\), \(v\in\{1,\dots,d_{\text{in}}\}\), $\gamma^{u,v}$ captures the dominant update direction at that coordinate: 
\begin{equation}
\gamma^{u,v} \;=\; \operatorname{sign}\!\Big(\sum_{c\in\mathcal{S}} \bm{\widehat{\delta}}_{c}^{\,u,v}\Big).
\label{eq:ties-sign-uv}
\end{equation}

Afterwards, we discard the contributions whose updates at  \((u,v)\) disagree with $\gamma^{u,v}$,  and average only the sign-consistent entries to produce the fused value at that coordinate:
\begin{align}
    \mathcal{A}^{u,v} &= \{c \in \mathcal{S}\, | \, \operatorname{sign}(\widehat{\bm{\delta}}_c^{u,v}) = \gamma^{u,v}\},\\
    \bm{\delta}_\text{merge}^{u,v} &= \frac{1}{|\mathcal{A}^{u,v}|} \sum_{c\in \mathcal{S}} \bm{\hat{\delta}}_c^{u,v}.
\end{align}

\begin{table*}[]
\centering
\resizebox{\textwidth}{!}{
\begin{tabular}{lcccccccccccccccc}
\toprule
\multirow{2}{*}{} &\multirow{2}{*}{Internalized?}& \multicolumn{5}{c}{Single -  Single} & \multicolumn{5}{c}{Single - Multiple} & \multicolumn{5}{c}{Multiple-Multiple}   \\ \cmidrule(l){3-7} \cmidrule(l){8-12} \cmidrule(l){13-17} 
&  & Acc & Prec. & Recall & F1 & bRecall&  Acc &  Prec. & Recall & F1 & bRecall &  Acc & Prec. & Recall & F1 & bRecall\\ \midrule
GPT4o + Prompt~\cite{hao2025rap}&\ding{56}&88.2&64.5&99.0&74.5&92.2&93.3&88.7&97.1&91.5&93.4&90.4&57.1&72.8&66.6&86.5\\
RAP-MLLM~\cite{hao2025rap}&\ding{56}& 85.3&59.7&98.1&71.6&90.5& 90.6& 84.8&94.4& 87.5& 91.8&87.4&52.3&68.8&57.4&80.4\\\hline
MyVLM~\cite{alaluf2024myvlm} &\ding{52} & 63.7&22.5&37.2&27.7&62.4&48.5&30.1&60.8&48.2&44.1&-&-&-&-&-\\
Yo'LLaVA~\cite{nguyen2024yo} & \ding{52}& 65.1 & 21.7 & 36.4 & 25.6 &58.2&  46.6 & 32.4 & 75.5 & 50.6 & 43.8 & 29.3 & 18.4 & 87.7 & 29.7 &52.2  \\
\rowcolor{gray!30}
Ours&\ding{52} & 73.6 & 31.0 & 86.4 & 44.1&  73.9  & 54.3 & 36.8 & 63.8 & 57.2 & 54.4 & 69.9 & 32.9 & 44.1 & 32.4 &  59.8  \\ \bottomrule
\end{tabular}}
\caption{Visual recognition performance under varying levels of compositional complexity: Single–Single, Single–Multiple, and Multiple–Multiple. Our approach achieves improvements over existing parameter-based approaches. Internalized indicates if each approach actually integrates the concept, \ie, concept representations are model dependent.}\label{tab:recog}
\end{table*}

For multi-concept inference, we apply the fused parameter to the base model, yielding the composed model: 
\begin{equation}
    \mathcal{M}_\text{comp} = \mathcal{M} \oplus \bm{\delta}_\text{merge}
\end{equation}
Our fusion operates entirely at inference with no extra training, so composition arises from internal parameter modularity rather than co-occurrence supervision.

\section{Experiments}

\subsection{Dataset} We construct our evaluation benchmark using images sourced from the recent multi-concept dataset~\cite{an2024mc}. The personalization pool consists of 118 distinct concepts, covering a diverse range of categories ranging from real-world individuals (\eg, celebrities and public figures), everyday objects, fictional or anime characters, \etc. For each concept, we use 10 positive images that depict that concept. To enhance discriminability and mitigate overfitting, we incorporate hard negative samples. For every concept, we retrieve approximately 100 visually similar images from CC12M~\cite{changpinyo2021conceptual} based on CLIP feature similarity\footnote{https://github.com/rom1504/clip-retrieval}. These visually close but semantically incorrect examples serve as challenging negatives during training. For evaluation, each concept is tested using its positive images along with 10 hard negatives collected using the same retrieval procedure. 

To assess zero-shot compositional generalization, we further evaluate on multi-concept images from~\cite{an2024mc}, where the model is not informed beforehand about which concepts appear in each image. This split includes 36 two-concept, 10 three-concept, and 2 four-concept combinations, enabling systematic analysis of multi-concept reasoning. Each combination has 5 images for testing.

\noindent\textbf{Training Data.}
We include four types of training data for each concept as follows:

\noindent\textit{- Positive Recognition}: Image query answer triples $(x, q, \ell)$ where $x$ contains the target concept and $q$ queries its presence (\eg, ``Can you see $\langle c\rangle$ in this photo?'') Answers affirm the concept (\eg, ``Yes, I can confirm that $\langle c\rangle$ is indeed in the photo.'')

\noindent\textit{- Hard Negative Recognition}: Triplets $(x^-, q, \ell^-)$ with visually similar but concept-absent images $x^-$, which we mine from the CC12M~\cite{changpinyo2021conceptual}. Prompts mirror positive case, while the answers negate the hypothesis (\eg, ``I've analyzed the image, and $\langle c\rangle$  is not present in the photo.'')

\noindent\textit{- Conversation}: We follow Yo'LLaVA~\cite{nguyen2024yo} and create conversation data for each concept.

\noindent\textit{- Image Captioning}: We prompt LLaVA~\cite{liu2023visual} to generate one caption for each concept image and use them for training. 

\subsection{Implementation Details} 

For training, we follow prior personalization work~\cite{nguyen2024yo} and adopt the LLaVA~\cite{liu2023visual} as the base vision-language model. We train the model with the AdamW~\cite{loshchilov2017decoupled} optimizer, employing a learning rate of $1e^-3$ for concept token and $1e^-4$ for LoRA parameters for 15 epochs. For each concept LoRA, we set the rank to be 4 and the scale to be 8. For sparsity ratio, we use $p=0.8$. For visual gating threshold $\tau$ in~\cref{eq:gating}, we set to 0.3. 

\noindent \textbf{Baselines.} 
We implement the following methods as our baselines to provide a comprehensive comparison against existing approaches. 

\noindent\textit{- MyVLM}~\cite{alaluf2024myvlm}: We use the same LLaVA backbone for a fair comparison and train each concept independently. However, since MyVLM~\cite{alaluf2024myvlm} operates with a single concept head at inference, extending it to handle multi-concept inputs is non-trivial and not directly supported by its architecture. 

\noindent \textit{- Yo'LLaVA}~\cite{nguyen2024yo}: We follow the original training protocol and learn each concept-specific prompt independently. During inference, the learned prompts are dynamically composed based on the same concept selection process. To enable multi-concept reasoning, we extend both the tokenizer and the classifier head to accommodate multiple personalized concept tokens.

\noindent \textit{- RAP-MLLM}~\cite{hao2025rap}: We construct the external database following the procedure described in \cite{hao2025rap}, providing for each concept an image and a  and employ the RAP-LLaVA model to obtain the corresponding results. As retrieval-based methods achieve personalization by referencing external concept images and textual information rather than embedding these concepts into the model itself. As a result, it shifts the burden of adaptation to the reasoning capabilities of the underlying VLM. 

\subsection{Visual Recognition}
We first evaluate our method on the task of visual recognition, where the goal is to correctly identify one or multiple concepts are present in an image. We consider three evaluation settings based on the relationship between the query and the visual content. \textbf{Single-Single} (SS) denotes the standard case, where the query corresponds to a single concept. \textbf{Single-Multiple} (SM), represents a more challenging setting, where the query targets a single concept but the image contains multiple co-occurring concepts.  Finally, \textbf{Multiple-multiple} (MM) evaluates the model's ability to jointly recognize multiple queried concepts that appear simultaneously in the image. 

For all three settings, we report accuracy, precision, recall, F1 score, and balanced recall (the mean of positive and negative recall), which together provide a comprehensive assessment of the model’s correctness, discrimination ability, and reliability under varying levels of visual complexity.

\cref{tab:recog} reports visual recognition performance across three evaluation regimes, and compares methods that rely purely on prompt (no internal concept learning) against models that explicitly internalized user concepts. As we can see, GPT4o when prompted similarly with RAP-MLLM~\cite{hao2025rap} achieves the highest overall performance on across all setting, surpassing the RAP-MLLM~\cite{hao2025rap}. This indicates that these methods are exceptionally good at detecting the presence of a queried concept when it truly appears. However, both model display a ``Yes-bias'', predicting concept presence even when absent. As a result, their precision drops sharply. %
For example, RAP-MLLM~\cite{hao2025rap} has a relatively low precision of 59.7\% compared to the high recall of 98.1\%. 

As for parameter-based models, we found that the precision scores are relatively low (20-30\% across settings), and the overall F1 performance remains weak even in the simplest SS task. Among these, our method achieves a more balanced performance profile, providing reliable recognition. Across all meetings, our model attains a higher precision and F1 than existing internalized approaches, demonstrating that our internalization is stronger and more stable. Notably, our balanced recall remains consistent (73.9\% / 54.4\% / 59.8\% across SS, MS, MM), showing that our approach is considerately more capable of rejecting false positives compared to  MyVLM~\cite{alaluf2024myvlm} and Yo'LLaVA~\cite{nguyen2024yo}. The gains are most visible in the most challenging MM setting, where our methods yields nearly 70\% accuracy, compared to that of Yo'LLaVA (29.3\%).

\begin{figure*}
    \centering
    \begin{overpic}[width=\linewidth]{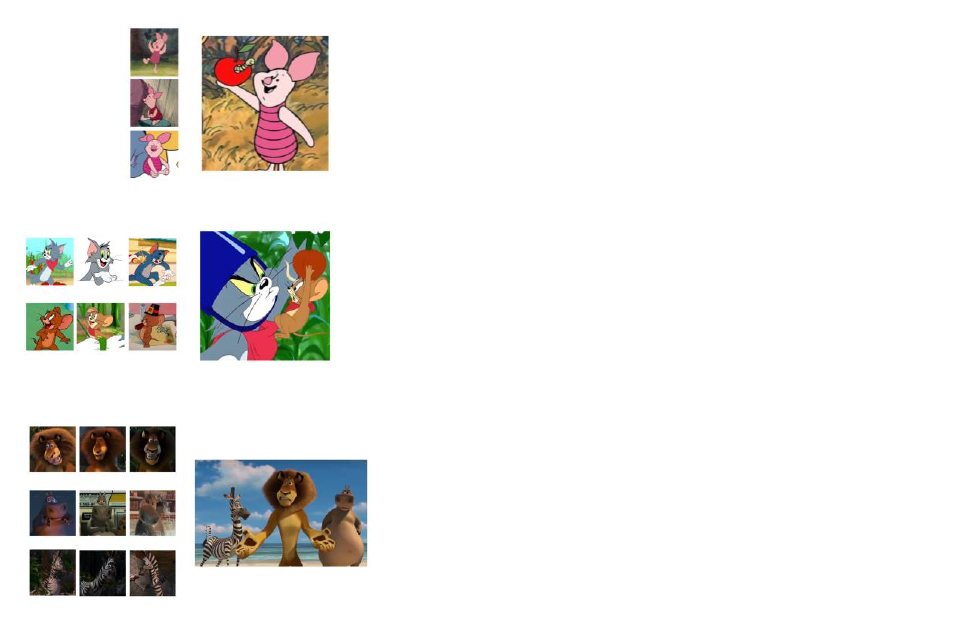}
        \put(6,64){\textbf{Concept}}
        \put(23,64){\textbf{Test image}}
        \put(56,64){\textbf{Generated personalized caption}}
        \put(9.5,53){\small \textlangle Z\textrangle }
        \put(0,36.5){\small \textlangle T\textrangle }
        \put(0,30.6){\small \textlangle J\textrangle }
        \put(0,17.8){\small \textlangle A\textrangle }
        \put(0,11){\small \textlangle G\textrangle }
        \put(0,5){\small \textlangle M\textrangle }

        \put(0,63){\line(1,0){100}} 
        \put(39.5,60){
            \parbox{0.6\linewidth}{ \textbf{RAP-MLLM}~\cite{hao2025rap}:
              \textlangle Z\textrangle\ is a pink pig with a red striped shirt. (\textcolor{red}{$^*$copy-paste}) 
            }
          }
        \put(39.5,54){
            \parbox{0.6\linewidth}{ \textbf{Yo'LLaVA}~\cite{nguyen2024yo}:
               In the image, we see the animated character \textlangle Z\textrangle, a pink-skinned pig with a joyful expression.  
            }
          }

        \put(39.5,47.5){
            \parbox{0.6\linewidth}{ \textbf{Ours}:
               In the image, we see the beloved character \textlangle Z\textrangle. She is standing in a field of tall grass. {In her right hand, she holds a red apple.}
            }
          }
        
        \put(0,43){\line(1,0){100}} 
         \put(39.5,39){
            \parbox{0.6\linewidth}{ \textbf{RAP-MLLM}~\cite{hao2025rap}:
             \textlangle J\textrangle\,  is being held by \textlangle T\textrangle.
            }
          }
        \put(39.5,34){
            \parbox{0.6\linewidth}{ \textbf{Yo'LLaVA}~\cite{nguyen2024yo}:
               In the image, we see two animated characters  \textlangle T\textrangle\ and  \textlangle J\textrangle.
            }
          }

        \put(39.5,27.9){
            \parbox{0.6\linewidth}{ \textbf{Ours}:
               In this image, we see \textlangle T\textrangle, a gray cat with a blue hat, is in hot pursuit of \textlangle J\textrangle, a brown mouse with a red scarf. 
            }
          }
         \put(0,23){\line(1,0){100}} 
          \put(39.5,19){
            \parbox{0.6\linewidth}{ \textbf{RAP-MLLM}~\cite{hao2025rap}:
              \textlangle A\textrangle, \textlangle M\textrangle, and \textcolor{red}{a hippo} stand together on a beach.
            }
          }
        \put(39.5,13){
            \parbox{0.6\linewidth}{ \textbf{Yo'LLaVA}~\cite{nguyen2024yo}:
               In the image, we see \textlangle M\textrangle\ is standing left on a sandy beach. In the center is \textcolor{red}{a lion} and on the right is \textcolor{red}{a hippopotamus}.
            }
          }

        \put(39.5,5.5){
            \parbox{0.6\linewidth}{ \textbf{Ours}:
               In the image, we see \textlangle M\textrangle\ is on the left. In the center is \textlangle A\textrangle. On the right is \textlangle G\textrangle. They are standing in a savanna-like environment with greenery and a clear sky in the background.
            }
          }
         \put(0,1){\line(1,0){100}} 
    \end{overpic}
    \caption{Qualitative comparison of personalized captioning capabilities across different models. From top to bottom, each row shows examples containing a single concept, two concepts, and three concepts, respectively. It shows our approach can be effectively extend to compositional scenarios. \textcolor{red}{Red} indicates the failure in generating the personalized concept.}
    \label{fig:captioning}
\end{figure*}

\subsection{Text / Visual QA}
\begin{table}[]
\centering
\begin{tabular}{@{}lcccc@{}}
\toprule
\multirow{2}{*}{} & \multicolumn{2}{c}{Text} & \multicolumn{2}{c}{Visual} \\ \cmidrule(l){2-5} 
 & Single & Multiple & Single & Multiple \\ \midrule
RAP-MLLM & 53.3 & 42.6 & 72.5 & 58.8 \\ \midrule
Yo'LLaVA & 71.5 & 57.3 & 64.7 & 52.4 \\
\rowcolor{gray!30}
Ours & 76.4 & 62.1 & 69.4 & 57.9 \\ \bottomrule
\end{tabular}
\caption{Performance on text-based QA and visual QA across single- and multiple-concept settings. Our method achieves the most balanced and robust results in both text-only and multimodal reasoning.}\label{tab:qa}
\end{table}
We first evaluate the model’s ability to perform concept-centric reasoning purely from text, without any visual evidence. This setting tests whether the model can reliably recall and interpret personalized concepts when prompted in isolation. As shown in Table \ref{tab:qa}, our method achieves 66.4\% accuracy in the single-concept setting and 52.1\% in the multi-concept setting, outperforming both RAP-MLLM (50.3\% / 34.6\%) and Yo’LLaVA (61.5\% / 47.3\%). For fair comparison, we also remove retrieved reference image in
RAP-MLLM~\cite{hao2025rap}, and it performs notably worse in this setup as the retrieved concept information may not help in answering detailed or fine-grained questions like ``What is the hair color of \textlangle c\textrangle?''.

We then extend the evaluation to the multimodal setting, where the model must ground concept-related questions in the visual content of the input image. This task is more challenging, as it requires aligning personalized concepts between text and vision. As shown in Table \ref{tab:qa}, our method attains 69.4\% accuracy in the single-concept VQA setting and delivers a competitive 57.9\% in the multi-concept scenario, demonstrating strong robustness as the visual reasoning complexity increases. Overall, our method provides the most balanced performance across both text and visual QA.

\subsection{Captioning}
Lastly, we evaluate the model on the task of captioning, which measures its ability to generate coherent and contextually accurate descriptions of images involving one or more learned concepts. We follow~\cite{alaluf2024myvlm} and report recall indicating whether the concept identifier appears at least once in the generated caption. 
Table \ref{tab:caption} reports personalized captioning recall in both the single-concept and multi-concept image settings. Across all methods, performance in the single-concept scenario is modest, and recall drops sharply once multiple concepts appear simultaneously. This pattern reflects the inherent difficulty of generating captions that accurately describe user-specific concepts.
RAP-MLLM~\cite{hao2025rap} obtains the highest single-concept recall at 60.0\%, outperforming internalized baselines such as Yo’LLaVA~\cite{nguyen2024yo} and our approach. However, closer inspection reveals that RAP-MLLM’s apparent advantage is largely driven by direct copying of the stored concept information. Specifically, we find that 77.4\% of captions containing the concept token are identical to the textual description stored in the concept database. This indicates that the model often relies on retrieval-based repetition rather than genuinely integrating the visual content. This limitation becomes even more apparent in the multiple-concept setting, where RAP-MLLM’s recall drops dramatically to 20.3\%. The reliance on verbatim retrieval prevents the model from resolving which personalized concept is actually present in the image, leading to mismatches and overly generic outputs.

In contrast, internalized methods such as Yo’LLaVA~\cite{nguyen2024yo} and our approach do not suffer from this copy-paste artifact. In the single concept setting, Yo'LLaVA achieves only 47.5\% and collapse to 7.6\% with multiple, highlighting the unstable concept retention and interference among token prompts when combined together. Our method improves recall in both settings (55.2\% and 12.8\%), demonstrating more robust concept grounding, but it too struggles when multiple personalized concepts must be composed in a single caption. The steep degradation in the multi-concept condition underscores that compositional personalized captioning remains a fundamentally challenging problem. 
\begin{table}[]
\resizebox{\linewidth}{!}{
\begin{tabular}{lcccc}
\toprule
 & Internalized? & Single & Multiple & Average \\ \midrule
RAP-MLLM~\cite{hao2025rap} & \ding{56} & 60.0 & 20.3 & 40.2 \\ \midrule
Yo'LLaVA~\cite{nguyen2024yo} & \ding{52} & 47.5 & 7.6 & 27.6 \\
\rowcolor{gray!30}
Ours & \ding{52} & 55.2 & 12.8 & 34.0 \\ \bottomrule
\end{tabular}}
\caption{Personalized captioning recall in single concept and multiple concept images.}\label{tab:caption}
\end{table}

We further present qualitative example to illustrate the personalized captioing behavior of different models in~\cref{fig:captioning}.  As we observed above, in the single-concept setting (first row), RAP-MLLM~\cite{hao2025rap} produces caption that directly repeat the concept information stored in the concept database. This behavior indicates a tendency to rely on memorized text rather than integrating visual evidence. In contrast, our method generates more concise and visually aligned descriptions that better reflect the content of the image.  In the multiple concept setting, our approach demonstrate stronger compositional generalization. While Yo'LLaVA~\cite{nguyen2024yo} occasionally fails to mention all relevant concepts, our method maintains more accurate coverage of the present concepts and produces captions that preserve both identity and context. These examples highlight the improved robustness of our fusion strategy in handling compositional personalization.

\subsection{Ablation Study}
\begin{table}[]
\centering
\begin{tabular}{@{}cccccc@{}}
\toprule
\multicolumn{1}{l}{$p$} & \multicolumn{1}{l}{0.3} & \multicolumn{1}{l}{0.5} & \multicolumn{1}{l}{0.7} & \multicolumn{1}{l}{0.8} & \multicolumn{1}{l}{0.9} \\ \midrule
Acc & 31.5 & 54.1 & 67.7 & \cellcolor{gray!30}{69.9} & 65.5 \\
F1 & 26.6 & 27.5 & 30.1 & \cellcolor{gray!30}{32.4} & 29.8 \\ \bottomrule
\end{tabular}
\caption{Ablation on the sparsity ratio $p$ on visual recognition task under the MM setting. }\label{tab:ablate}
\end{table}
\textbf{Sparsity ratio $p$.} \cref{tab:ablate} reports how varying the sparsity ratio in the fusion process influences the visual recognition performance on the MM setting. We observe a clear trend that increasing the sparsity ratio from 0.3 to 0.7 leads to substantial gains in both accuracy (31.5\% vs. 67.7\%) and F1 score (26.1 vs 30.1). Performance peaks around 0.7–0.8 sparsity, where the model achieves the best accuracy (67.7–69.9) and the highest F1 (30.1–32.4). This indicates that moderate sparsity helps suppress interference among LoRA updates, allowing the fused model to better isolate concept-specific signals and handle compositional scenes. 
Beyond this point, however, increasing sparsity to 0.9 causes a noticeable drop in both metrics. It is likely that much of the meaningful adaptation signal has been removed. 

\begin{table}[]
\centering
\begin{tabular}{@{}lccc@{}}
\toprule
\multirow{2}{*}{$\tau$} & \multicolumn{2}{c}{Visual Rec.} & Captioning \\ \cmidrule(l){2-4} 
 & Acc & F1 & Recall \\ \midrule
0.1 & 68.1 & 30.9 & 10.6 \\
\rowcolor{gray!30}
0.3 & 69.9 & 32.4 & 12.8 \\
0.5 & 69.9 & 32.4 & 8.2 \\ \bottomrule
\end{tabular}
\caption{Ablation study on the similarity threshold $\tau$ on visual recognition (MM) and captioning (Multiple).}\label{tab:tau}
\end{table}
\noindent\textbf{Visual gating threshold $\tau$.} \cref{tab:tau} summarizes the impact of the visual gating threshold $\tau$ on multi-concept performance. For visual recognition (MM), both accuracy and F1 improve when increasing $\tau$ from 0.1 to 0.3, indicating that a modest threshold helps suppress irrelevant LoRA activations. Performance remains stable at $\tau=0.5$, largely because the gating mechanism also incorporates textual cues, which reliably activate the correct concept module even when the visual gate becomes more selective. As a result, recognition does not degrade despite the stricter visual threshold. In contrast, captioning is more sensitive to the choice of $\tau$. Concept recall peaks at $\tau= 0.3$, whereas both lower (0.1) and higher (0.5) thresholds lead to substantial drops. A low threshold can activate too many irrelevant LoRAs, introducing interference, while a high threshold may suppress valid concepts that lack of strong visual similarity. Unlike recognition, which benefits from textual gating, captioning depends more heavily on visual cues, making it more sensitive to $\tau$.

\subsection{Limitations}
Although our method improves compositional personalization, several limitations remain. One is that our approach assumes that each personalized concept can be cleanly captured by its own token and LoRA module; highly entangled or visually ambiguous concepts may require more expressive adaptation mechanisms. Another limitation is that while the sparsification strategy enhances modularity and reduces interference, it also poses the risk of losing fine-grained details or highly variable appearances of concepts. 

\section{Conclusion}
We presented Gate-and-Merge, a framework for zero-shot compositional personalization of VLMs that learns each user-defined concept independently and composes them at inference without co-occurrence supervision. Our method represents each concept using a lightweight token and a LoRA module attached to the language branch, while the vision encoder remains entirely frozen for efficiency and stability. At test time, a multi-modal gating mechanism identifies the concepts relevant to the input, after which their corresponding LoRA updates are selectively fused in weight space to form a single compositional adapter. On multiple personalization tasks, our approach demonstrates substantial improvements in both single-concept and multi-concept accuracy. 
{
    \small
    \bibliographystyle{ieeenat_fullname}
    \bibliography{main}
}

\end{document}